\documentclass[runningheads]{llncs}
\usepackage{graphicx}
\usepackage{cite} % Load the cite package for handling citations
\usepackage{enumitem}
\usepackage{siunitx}
\usepackage{ulem}
\usepackage[table, xcdraw]{xcolor}
\usepackage{diagbox}
\usepackage{multirow}
\usepackage{booktabs}
\usepackage{mathrsfs}
\usepackage{amsmath}
\usepackage{amsfonts}
\usepackage{makecell}
\usepackage{hyperref}
\hypersetup{
    colorlinks=true, 
    linkcolor=blue, 
    filecolor=magenta,     
    urlcolor=cyan, 
    pdftitle={Overleaf Example}, 
    pdfpagemode=FullScreen, 
    }
\begin{document}

\title{The MRI Scanner as a Diagnostic: 
\\Image-less Active Sampling}
\titlerunning{The MRI Scanner as a Diagnostic: Image-less Active Sampling}
\author{
Yuning Du\inst{1}\and Rohan Dharmakumar\inst{2} \and Sotirios A.Tsaftaris \inst{1, 3}}
\index{Du, Yuning \and Dharmakumar, Rohan\and  }

\authorrunning{Y.Du et al.}
\institute{School of Engineering, The University of Edinburgh, Edinburgh, EH9 3FG, UK \and Krannert Cardiovascular Research Center, Indiana University School of Medicine, Indianapolis, Indiana, USA\and The Alan Turing Institute, London, NW1 2DB, UK\\}
% \author{Anonymous Authors}
% \authorrunning{Anonymous et al.}
% \institute{Anonymous Institution}
%
\maketitle              % typeset the header of the contribution
\begin{abstract}
Despite the high diagnostic accuracy of Magnetic Resonance Imaging (MRI), using MRI as a Point-of-Care (POC) disease identification tool poses significant accessibility challenges due to the use of high magnetic field strength and lengthy acquisition times.
We ask a simple question: \textit{Can we dynamically optimise acquired samples, at the patient level, according to an (automated) downstream decision task, while discounting image reconstruction?}
We propose an ML-based framework that learns an active sampling strategy, via reinforcement learning, at a \textit{patient-level} to directly infer disease from undersampled $k$-space. We validate our approach by inferring Meniscus Tear in undersampled knee MRI data, where we achieve diagnostic performance comparable with ML-based diagnosis, using fully sampled $k$-space data. We analyse task-specific sampling policies, showcasing the adaptability of our active sampling approach. The introduced frugal sampling strategies have the potential to reduce high field strength requirements that in turn strengthen the viability of MRI-based POC disease identification and associated preliminary screening tools.

\end{abstract}
\section{Introduction}
\label{sec:Introduction}
Despite the proliferation of Magnetic Resonance Imaging (MRI), its role as a Point-Of-Care (POC) diagnostic tool is muted due to poor accessibility caused by long acquisition time and cumbersome equipment~\cite{geethanath2019accessible}. Consequently, advancements have been made in assisting the diagnostic process with machine learning methods to sample less $k$-space data and reduce acquisition time in turn~\cite{lin2021artificial}. 
%However, the need for high-fidelity images that radiologists can accurately interpret puts lots of constraints when implementing the machine learning method in the diagnostic process. Therefore, we need to reconsider the MRI-based diagnostic process. 

\begin{figure}[t]
\begin{center}
\includegraphics[width=0.8\textwidth]{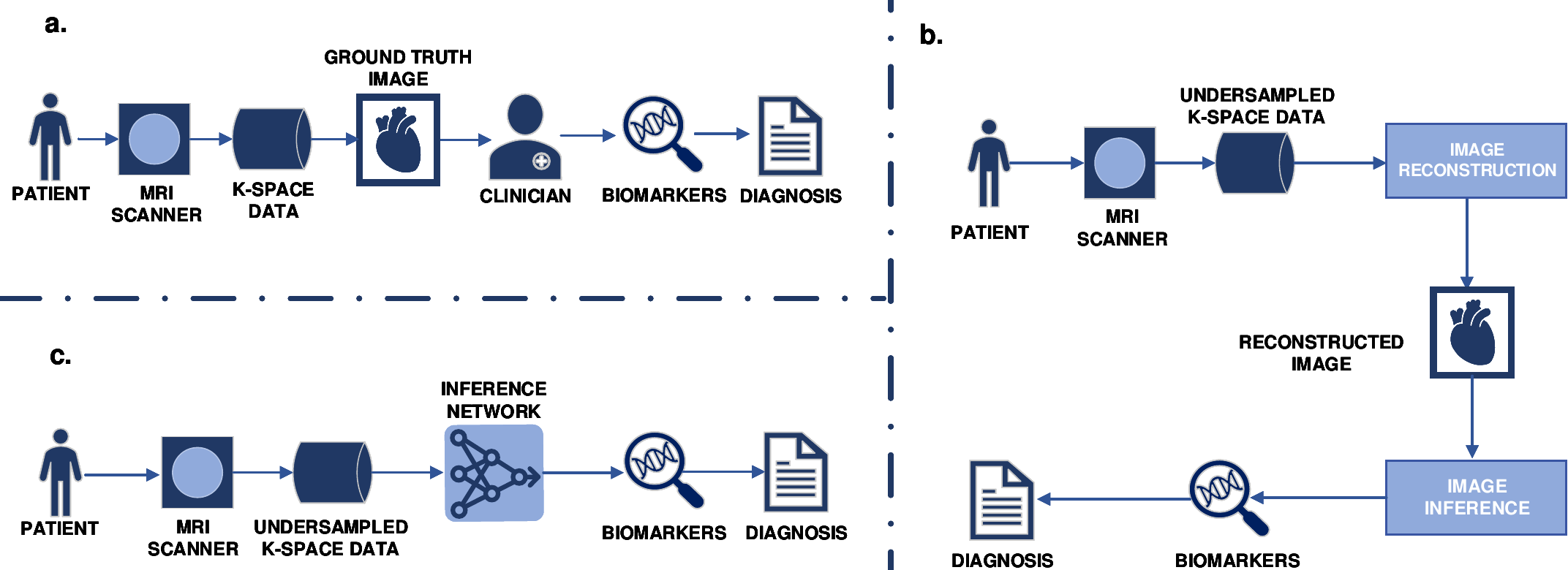}
\end{center}
\caption{Conventional vs. ML-based Diagnostic Processes.} \label{fig1}
\vspace{-1mm}\end{figure}

We first illustrate how learning-based strategies can help to reduce acquisition time by considering the depiction of conventional MRI-based diagnostic processes. In Figure~\ref{fig1}a. an MRI scanner samples the full $k$-space, resulting in high-fidelity images, which are further interpreted by professional radiologists to identify biomarkers and provide a diagnosis. Alternatively, machine-learning (ML) based diagnostic processes can enable acquisition time savings by reconstructing a high-fidelity image from undersampled $k$-space (Figure~\ref{fig1}b). Previous work has focused on optimising both reconstruction~\cite{cai2020review} and $k$-space sampling strategies~\cite{bahadir2020deep}. To find the best $k$-space sampling pattern, various methods~\cite{wang_joint_2022,xuan_learning_2020,ravula_optimizing_2023} optimise the mask given pre-defined sample rates, resulting in \textit{population-level} masks, while reinforcement-learning-based methods~\cite{bakker_experimental_2020,pineda_active_2020} can optimise an active sampling strategy at population or patient-level. A re-interpretation of such conventional use of MRI can lead to considerable savings, which will largely enable a low-field MRI future opening the road to POC and bedside imaging.
%
%Furthermore, ML-based image classification and segmentation~\cite{cai2020review} can provide image interpretation assistance.
%

%
%The conventional process ensures high-quality diagnosis but is time-consuming, while ML-based diagnostic processes tend to optimize each stage individually.  

%ML-based processes can offer a compelling alternative. 
%acquisition time through reconstruction and labor costs through image analysis. 
%However, a naive view implies that we directly combine image reconstruction with analysis methods.
%, spurious correlations may be introduced in the diagnostic process due to the reconstructed $k$ space. 
Considering that image reconstruction is a shared stage in both conventional (Figure~\ref{fig1}a) and ML-based (Figure~\ref{fig1}b) diagnostic processes, one ponders \textit{Is the image necessary for automated, diagnostic, inferences?} %An ideal AI agent 
One option is to perform direct inference on undersampled $k$-space as shown in Figure~\ref{fig1}c. In fact, \cite{schlemper2018cardiac} have shown that it is possible to obtain biomarkers directly from $k$-space data.  However, one then immediately ponders \textit{Can AI agents learn effective $k$-space sampling strategies by considering expected diagnostic performance? }
%thus optimising the sampling cost of the diagnostic process? 
%This can be the initial diagnosis to screen high-risk patients in need of urgent intervention to effectively save diagnosis time at population-level. 
A recent realisation of this concept is presented in~\cite{singhal_feasibility_2023}. The work formulates the described process as a classification task with an optimised sampling strategy. However, the sampling strategy is designed at the \textit{population-level}, resulting in the same mask for all patients.
%In such case, an active sampling strategy based on reinforcement learning can be an accessible approach for a customised sampling mask at the patient level. 
All approaches that find optimal patient-level active sampling strategies use image reconstruction~\cite{bakker_experimental_2020,pineda_active_2020,zhang_reducing_2019}. To the best of our knowledge, active sampling strategies for patient-level disease inference, from undersampled $k$-space, are lacking. Motivated by this, our \textbf{contributions} include:
\begin{enumerate}
    \item[$\bullet$] We propose, the first, patient-level active sampling using a reinforcement learning framework, aiming for direct disease inference from $k$-space. 
    %MRI-based diagnosis framework for disease inference without high-fidelity images, aiming for initial POC diagnosis.
    \item[$\bullet$] We test feasibility to infer Meniscus Tear presence in undersampled MRI.
    %and show that a policy can be trained to optimise $k$-space acquisition leading to savings in time.
    %and our method achieves competitive performance to the \textit{Oracle} Method with 25\% data usage, and $k$-space savings compared to the patient-level benchmark without fully sampled $k$-space.
    \item[$\bullet$] We investigate how different policies make decisions and how different starting points alter behaviour.
\end{enumerate}

\section{MRI Undersampling Preliminaries} 
Instead of directly imaging the human anatomy, MRI captures the electromagnetic activity in the body after exposure to magnetic fields and radiofrequency pulses, which can be measured in $k$-space (i.e., the frequency domain). Considering the single coil measurement, the $k$-space data can be represented as a 2-dimensional complex-valued matrix $x \in \mathbb{C}^{r \times c}$, where $r$ is the number of rows and $c$ is the number of columns. The spatial image $I$ can be obtained by applying the inverse Fourier Transform to $x$, denoted as $I = \mathcal{F}^{-1}(x)$. The undersampled $k$-space can be therefore represented as $x_s= U_L\circ x$, where $U_L$ can be viewed as a binary mask $U \in \{0, 1\}^{r \times c}$ with L measurements from $k$-space. In our work, we exclusively consider the Cartesian mask for MRI undersampling.  Consequently, the undersampled image is denoted as $I_s = \mathcal{F}^{-1}(x_s)$.
\begin{figure}[t]

\begin{center}
\includegraphics[width=0.8\textwidth]{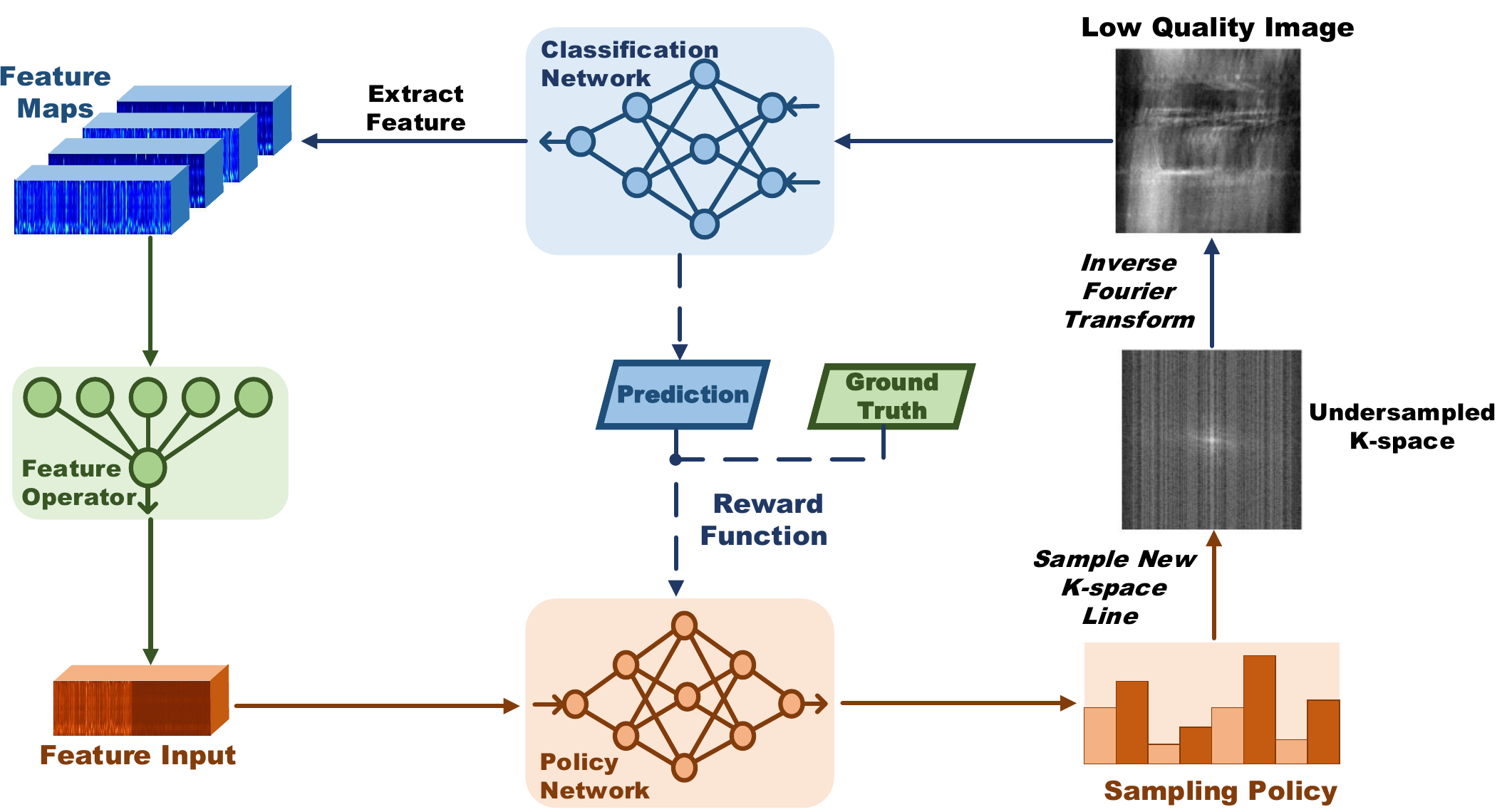}
\end{center}

\caption{Our framework for active MRI sampling for point-of-care diagnosis.} \label{fig2}

\end{figure}

\section{Methods}
\label{method}
\noindent \textbf{Overview}:
Our framework, illustrated in Figure~\ref{fig2}, aims to reduce acquisition time by selectively sampling the $k$-space according to diagnostic significance, in a progressive fashion. The process begins with a small subset of randomly sampled $k$-space with L measurements denoted as \(x_{s_0} = U_L\circ  x\), resulting in a low-quality image \(I_{s_0}\). \(I_{s_0}\) is then input into a pre-trained classification network \(f\) to generate the initial prediction \(y_0\). The extracted feature maps \(m_0\) are processed by the \textit{Feature Operator} \(O\), producing high-level features \(h_0 = O(m_0)\), which are then fed into the \textit{Active Sampler} \(S\). The policy network generates a sampling policy \(p_\phi(h_0)\) parameterised by $\phi$, guiding the selective sampling of diagnostically significant lines from the $k$-space. These sampled lines are subsequently added to \(U_L\). The updated undersampled $k$-space data at step \(t \in [0, T]\) are denoted as \(x_{s_t} = U_{t}\circ  x\), and are then fed back into the classification network with inverse Fourier transform. The iterative process continues until a sampling budget \(T\) is exhausted or user-defined reliability criteria are satisfied. 
During the training stage, the predictions \(y_t\), accompanied by the ground truth diagnostic label \(g\), are used as the criterion \(r_t = r(y_t, g)\) to supervise the $S$. While in inferece stage,  there is no ground truth diagnostic label \(g\) provided.

\noindent \textbf{Classification Network and Image Operator}: 
To improve stability,
%introducing additional variance into the framework, 
the classification network is pre-trained with undersampled $k$-space data, ensuring an advantageous reward for the active sampler during training. This network also functions as the feature extractor in our setting. In the classification network, the earlier layers tend to learn low-level features such as edges, textures, and simple patterns, while deeper layers learn more complex and high-level features that are useful for discriminating between different classes. Therefore, utilising the feature maps from the model, the high-level features are selected and processed by the feature operator as further inputs.

\noindent \textbf{Active $k$-space Sampler with Greedy Policy}:
Inspired by~\cite{bakker_experimental_2020,pineda_active_2020}, the sequential selection of $k$-space can be formalised as a Partially Observable Markov Decision Process (POMDP)~\cite{sondik1971optimal}. Greedy Policy is used to maximise the expected return \(J(\phi)\) of a policy \(p_\phi\) parameterised by \(\phi\) in such a POMDP. At each step, the classification improvement can be calculated using \(R(r_t, r_{t+1}) = r(y_{t+1}, g) - r(y_t, g)\), where the criterion \(r\) is the cross-entropy.

During inference, the agent will be sampling one-line at a time. However, this will slow during training. Hence, the policy network is trained by sampling several lines in parallel the rewards of which are averaged~\cite{kool2019buy}. Formally, we sample 
\(q\) lines at every time step, for a reward \(R_{i, t}\) as the reward obtained from sample \(i\) at time step \(t\), to obtain the following estimator:
\begin{equation}\label{eq1}
\resizebox{0.9\linewidth}{!}{$
\nabla_\phi J(\phi) \approx \frac{1}{q-1} \mathbb{E}_{x} \sum_{i=1}^q \sum_{t=L}^{T-1} \left[\nabla_\phi \log p_\phi\left(h_t\right)\left(R_{i, t} - \frac{1}{q} \sum_{j=1}^q R_{j, t}\right)\right].
$}
\end{equation}

\noindent \textbf{Evaluation Metrics}:
To assess the network's classification performance, we employ metrics such as Recall, Area Under Curve (AUC), and Specificity. 
% ---- Experiments ----

%+++

\section{Experiments}
\label{sec:Experimental Analysis}
\subsection{Dataset and Pre-processing}
\label{sec:Data}
\noindent \textbf{Dataset}:
We used single-coil $k$-space data and slice-level labels from the publicly available fastMRI dataset~\cite{zbontar2018fastmri} and fastMRI+ dataset~\cite{zhao2021fastmri+}. Randomly selecting $1100$ annotated volumes ($39125$ slices) from the fastMRI Knee dataset. Our diagnostic task is to identification  Meniscus Tear (MT)  in each slice. Thereby, there are $32035$ train slices ($11.8\%$ with MT), $5249$ validation slices ($9.3\%$ with MT), and $1841$ test slices ($11.1\%$ with MT).
\begin{table}
\centering
\caption{Diagnosis Support and Data Access of Implemented Methods.}
\label{tab1}
\resizebox{0.9\linewidth}{!}{
\begin{tabular}{|c |c |c |c |c |c|}     
\hline
 & \multicolumn{2}{c|}{Diagnostic Support} & \multicolumn{3}{c|}{Data Used by the Model} \\
 \hline
Method& \makecell{Sampling\\Optimization} & \makecell{Patient-level\\ Strategy} & \makecell{Full $k$-space} & \makecell{Diagnostic Label} & \makecell{Undersampled $k$-space} \\
\hline
\textit{Oracle} &  &  & \checkmark & \checkmark &  \\
\textit{Undersampled} &  &  &  &  \checkmark&\checkmark\\
\textit{Policy Reconstruction}~\cite{bakker_experimental_2020}& \checkmark& \checkmark & \checkmark & \checkmark &\checkmark  \\
\textit{Policy Classifier}\textbf{(Ours)} &  \checkmark&  \checkmark&  &\checkmark  &\checkmark  \\
\hline
\end{tabular}}
\end{table}

\noindent \textbf{Data Pre-processing}:
Since the $k$-space data have various sizes, we first use inverse Fourier transform to the fully sampled $k$-space data  to get the ground truth image and crop it to size $(320 \times 320)$ for computation convenience. Thereby, the fully sampled $k$-space data of uniform size can be obtained by applying Fourier transform to the ground truth image. 
Notably, there is a severe class imbalance regarding the MT identification task. During training the classification network, we oversample the data to avoid overfitting on the majority class and poor generalisation on the minority class. 

\subsection{Implementation Details}
We devise inference benchmarks to allow us to evaluate our approach in a fair fashion. Two benchmarks have the access to fully sampled $k$-space(high-fidelity images), and two do not optimise at the patient level the sampling strategy. The diagnosis support and data access are shown in Table~\ref{tab1}.
%population level 
%We highlight a key difference that two of our benchmarks employ: access to high fidelity, fully sampled data with annotations. Our approach does not assume this. 

\noindent \textbf{Fully Sampled (\textit{Oracle}):} This serves as an benchmark estimator of classifier performance on image input obtained by fully sampled $k$-space data, and hence no sampling optimisation occurs.  We trained the classifier with the ground truth image as input which is transformed from fully sampled $k$-space data and supervised by the diagnostic label using cross-entropy loss. We use as classification backbone a ResNet-50~\cite{he2016identity} and to address the class imbalance in the training set, we add extra dropout layers overfitting, resulting in a total of $25.6 \mathrm{M}$ parameters. 
%This is also the population-level benchmark.

\noindent \textbf{Undersampled}: This classifier serves as a baseline of performance when simple inverse Fourier is used to transform the under-sampled $k$-space data, without any sampling optimisation. It has the same backbone as the \textit{Oracle}, and is trained with undersampled images with various sample rates ($4$ to $20$) and center fraction ($0$ to $0.10$), supervised as before. 
%diagnostic label using cross-entropy loss, referred to as the Undersampled Image Classifier (\textit{Undersampled}). It serves as the baseline for our method.

\noindent \textbf{Policy (via) Reconstruction~\cite{bakker_experimental_2020}:}  We compare with a model that optimises patient-level sampling strategy with image reconstruction error as rewards. Notably this method has access to fully sampled $k$-space, and hence has access to more information during training. 
We pre-trained a reconstruction network using a U-Net~\cite{ronneberger2015u} as the backbone with a first feature map size of $16$ and $4$ pooling cascades, resulting in a total of $837 \mathrm{K}$ parameters. The reconstruction network is trained with various sample rates ($4$ to $20$) and center fraction ($0$ to $0.10$) supervised using the $\ell_1$ loss. 
We train the active sampler for reconstruction with reconstructed images from pre-trained model as inputs, and Structural Similarity Index Measure as the criterion to provide rewards, resulting in a total of $26.7 \mathrm{M}$ parameters. The reconstructed images obtained with the policy network and pre-trained reconstruction model are evaluated with the \textit{Oracle}, referred to as the Policy-based Reconstruction Network (\textit{Policy Reconstruction}).

\noindent \textbf{Proposed Policy Classifier}:
For our method, a backbone similar to the \textit{Undersampled} is used as pre-trained classification network. The reward is driven by the predictions of the network and its feature maps are used to train the policy. The Feature Operator uses the last 2 layers' feature maps as input and the global average pooling function to achieve a $(80, 512)$ output to feed in the policy network.  
The policy network ($11 \mathrm{M}$ parameters) uses the cross entropy from classification network to provide rewards. It is trained with an initial sample rate of $5\%$ with multiple center fraction and samples $64$ lines to reach the sample rate of $25\%$ as the sampling budget. The parallel acquisition $q=8$. 
%Our framework is referred to as Policy-based Undersampled classifier (\textit{Policy Classifier}).

For all methods, we employ the Adam optimiser with a learning rate of $10^{-4}$ and a step-based scheduler with a decay gamma of $0.1$ for all model training. All classification and reconstruction models are trained for $30$ epochs, and the policy networks of the active sampler are trained for $20$ epochs. Our experimental setup uses the PyTorch framework, and all computations are conducted on NVIDIA A100 Tensor Core GPUs.
Our code\footnote{https://anonymous.4open.science/r/KspaceToDiagnosis-CB5E}  is available.

\iffalse 

\begin{table}[t]
    \centering
    \caption{Image Classification Performance for Different Methods. Numbers in brackets indicate classifiers that closely match the performance of the \textit{Oracle}. Numbers in bold indicate the best performance.}
    \resizebox{0.8\linewidth}{!}{%
        \begin{tabular}{c|c|c|c}
        \hline
        \textbf{Model} & \textbf{AUC$\uparrow$} & \textbf{RECALL$\uparrow$} & \textbf{SPECIFICITY$\uparrow$} \\
        \hline
        \textbf{Oracle}& 0.819 (0) & 0.816 (0) & 0.822 (0) \\
        \textbf{Undersampled}& 0.756 (-0.063)& 0.694 (-0.121) & 0.817 (-0.005)\\
        % \textbf{Recon Classifier} & 0.687 (-0.132) & \textbf{0.402 (0.014)} & 0.510 (-0.306) & 0.864 (0.042) \\
        \textbf{Policy Reconstruction}& \textbf{0.845 (0.027)}& \textbf{0.908 (0.092)} & 0.783 (-0.039) \\
        \textbf{Policy Classifier}& 0.794 (-0.025)& 0.743 (-0.073)& \textbf{0.845 (0.023)} \\
        \hline
        \end{tabular}}
        \label{table1}
\end{table}

\fi

\subsection{Results}
\begin{figure}[t]
    \centering
    \includegraphics[width=0.9\linewidth]{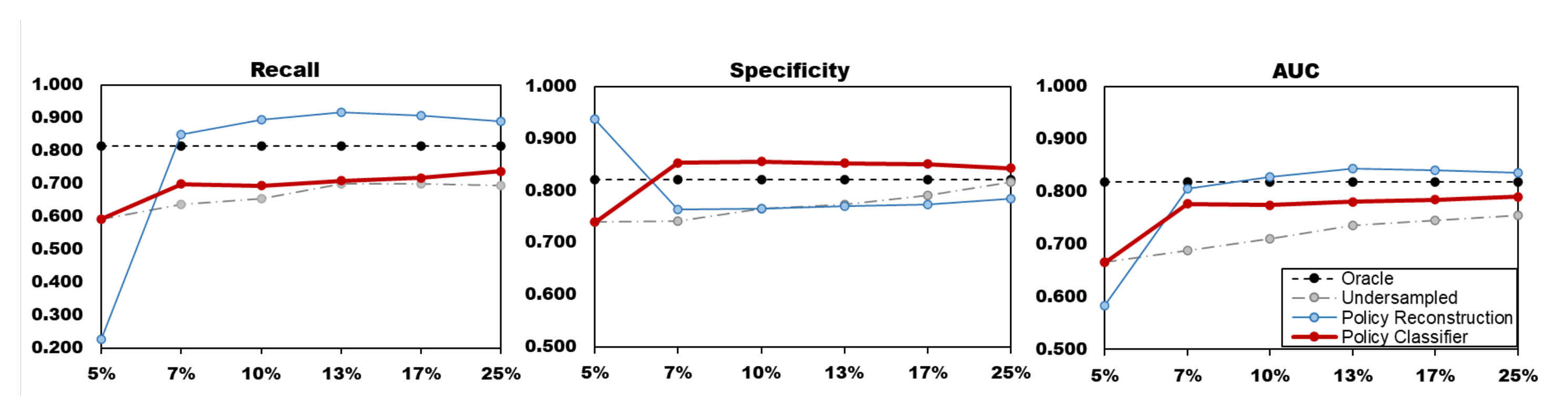}
    \caption{Classification performance varying sample rates (horizontal axis). More metrics are reported in the Supplementary.}
    \label{fig3}
\end{figure}
%
% The result of the US Classifier represents the lower bound\edward{bound for what quantity?} since it predicts with limited and random-selected $k$-space data, indicating minimum performance achievable with the undersampled data. The Oracle represents the upper bound as it conducts identification given access to the fully sampled $k$-space information. Thus, an ideal classifier should exceed the performance of the lower bound and try to achieve similar performance as the upper bound.
% \edward{agree with Sotos, very unclear}

\noindent \textbf{Making Diagnostic Decisions with Undersampled Data} \\
%Table~\ref{table1} presents the image classification performance when exhausting the sampling budget, with $25\%$ of the data sampled from $k$-space.Given fixed sample rate, when compared to population-level baseline \textit{Undersampled}, \textit{Policy Classifier} provide better performance in all three metrics. Our method also closely matches the AUC and recall of population-level benchmark $Oracle$, with values of $0.794$ and $0.743$, respectively. Meanwhile, our method outperforms \textit{Oracle} and \textit{Policy Reconstruction}  in specificity with the value of $0.845$.  
Figure~\ref{fig3} compares the performance at various sample rates for three strategies that use undersampled masks, namely the \textit{Undersampled},  \textit{Policy Reconstruction} and \textit{Policy Classifier}. The first randomly samples $k$-space lines progressively; the other two start with a randomly initialised mask with a sample rate of $5\%$ and optimally decide using their respective rewards and policies (see Section~\ref{method}).
%in the \textit{Policy Reconstruction} and \textit{Policy Classifier}. 
Our method, the \textit{Policy Classifier}, consistently outperforms the \textit{Undersampled} baseline across all metrics, showing that optimal sampling via the policy network helps in identifying diagnostically relevant $k$-space lines. This leads to considerable savings in data sampled (and consequently scan time). Our model approximates well the performance of the \textit{Oracle}, which has been trained on high-fidelity data, and reaches optimal performance quickly. Taking AUC as an example, our approach reaches an AUC of 0.780 with 7\% of the sampled $k$-space.  
%our method progressively approaches its performance and even outperforms it in specificity using less than 25\% $k$-space data. 
Our method closely approximates the performance of the \textit{Policy Reconstruction}, which we highlight has been trained with access to fully sampled $k$-space.\footnote{ We observe that the \textit{Policy Reconstruction} performs better than the \textit{Oracle}. This policy reconstructs an image which is given to the \textit{Oracle} for classification. Some denoising and smoothing are happening at the reconstruction which in turn acts as a regulariser for the \textit{Oracle} classifier explaining this slightly improved performance. } 
%In the discussion we mention mechanisms to  denoising aspect of the netow 
%our method approaches the performance of the Oracle in the same speed as the \textit{Policy Reconstruction} in recall and AUC, and outperforms it in specificity after sampling only 7\% of $k$-space data.  

\begin{figure}[t]
    \centering
    \includegraphics[width=0.9\linewidth]{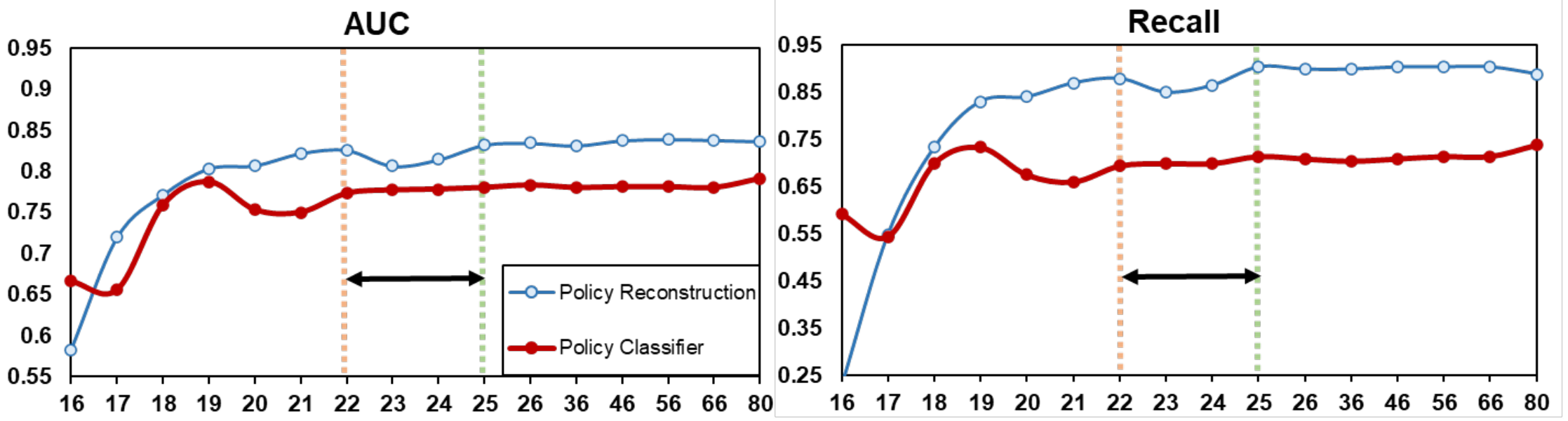}

    \caption{$k$-space behaviour of the two policies. The horizontal axis indicates the cumulative lines acquired while sampling.The initial 16 lines are randomly sampled.}
    \label{fig4}

\end{figure}

\noindent \textbf{$k$-space Sampling Behaviour of the Policies}\\
The results of the previous paragraph are reported over a coarse percentage of $k$-space lines acquired. It is worth looking into how the two different policies behave when asking each policy to progressively make decisions on which line to acquire in a line-by-line manner. The results of this exercise are shown in Figure~\ref{fig4}. % with x-axis showing cumulative lines acquired (both starting with 16 random lines in the center).
%during the progressive sampling of  $k$-space by \textit{Policy Reconstruction} and \textit{Policy Classifier},  after the sample rate reaches 7\%, the performance tends to be stable and the increase in performance becomes less intensive. To find the stable point of the performance during sampling, we evaluate the diagnostic performance in a stepwise manner as shown in Figure~\ref{fig4}. Compared to \textit{Policy Reconstruction}, our method owns earlier stable points in recall and AUC with 3 steps less, and has the same stable point as \textit{Policy Reconstruction}. The difference in stable point represents the $k$-space savings, indicating our method locates the diagnosis relevant $k$-space more efficiently than \textit{Policy Reconstruction} even under the fact that our method doesn't have access to fully sampled $k$-space as \textit{Policy Reconstruction}.
The behaviour of the two methods is evidently different. Taking Recall and AUC as examples, the \textit{Policy Classifier} plateaus quickly, reaching excellent performance with 22 sampled lines and continuing to make, small, improvements. The \textit{Policy Reconstruction} over the same interval makes sub-optimal (to classification) decisions before it plateaus. This behaviour can be explained by the fact that a reconstruction policy may not favour lines useful for classification. The \textit{Policy Classifier} appears to change behaviour between 19 and 21 lines acquired. This can be attributed to a drift from how the classifier was pre-trained with randomly sampled lines. We discuss solutions to this later.

\begin{figure}[t]
\begin{center}
\includegraphics[width=0.8\textwidth]{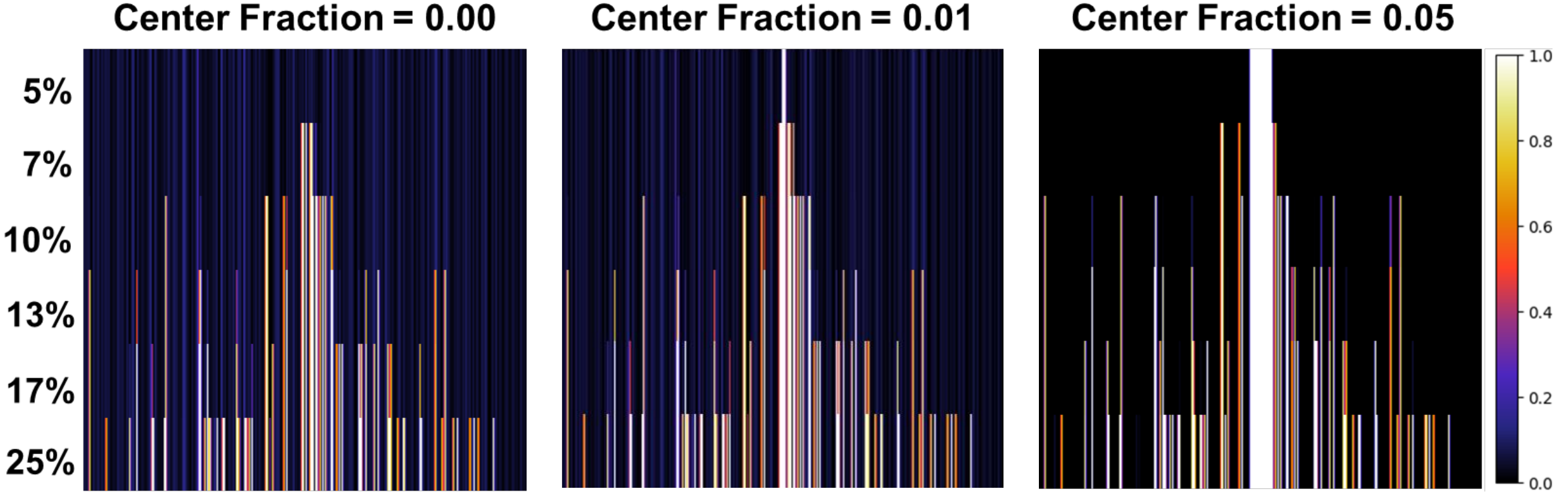}
\end{center}
\caption{$k$-space preference as a function of sample rate (vertical axis downward) and for 3 center fraction scenarios. Colorbar indicates the possibility of being sampled. } 
\label{fig5}
\end{figure}

\noindent \textbf{Task-specific `Coarse-to-Fine' Sampling Policy}\\
It would be interesting to see which $k$-space lines are favoured by our policy and how altering the starting point (namely different percentage of center fraction which represents the amount of low-frequency $k$-space intentionally sampled from the center of the $k$-space) modulates this behaviour.  Figure~\ref{fig5} illustrates the average masks provided by three policy networks.   
Comprared with the sampling policy for reconstruction in supplementary,  this analysis reveals that early on the policy favours both low-frequency and some high-frequency data. However, later on the policy favours specific  diagnostically-relevant $k$-space lines of high-frequency. Hence we see the policy capturing `Coarse' features such as anatomical structures or essential patterns first and later `Fine' details from the high-frequency data. 
When center fraction increases, the policy samples even more high-frequency information early on. While this might be obvious it is actually driven by low AUC performance when forcing the model to sample more center lines from the start (the AUC drops from $0.794$ to $0.743$ when the center fraction is set to $0.05$, see also more metrics in the supplementary). By forcing the model to sample redundant low-frequency data, the policy tries to recover to identify diagnostically significant data within the $k$-space.

\section{Discussion} 

%\noindent \textbf{Performance of Policy Reconstruction}:
%Notably, the Policy Reconstruction outperforms our method and achieves results comparable to the upper bound when evaluated solely based on metrics. However, as mentioned in Section~\ref{sec:Introduction}, we cannot guarantee that the performance improvement is solely due to the reconstructed $k$-space having less confounding information or if it introduces spurious correlations into the classification process. In contrast, our method relies purely on data collected from the ground truth $k$-space, delivering comparable results to the upper bound and effectively reducing acquisition time.    

\noindent \textbf{Clinical Relevance}:
Our method actively samples $k$-space data whilst simultaneously conducting disease inference, enabling real-time diagnosis during scanning. This approach eliminates the need for sampling full $k$-space,
%and the immediate involvement of a radiologist for image interpretation, 
thereby reducing acquisition time. In addition, our approach does not require high-fidelity fully sampled data for policy training, which is a hard requirement of reconstruction policies. %This effectively opens the road to the 
We believe this has the potential to further accelerate the development of low-field MR. %Critically our method provides patient-level sampling strategies toward time optimisation that can ultimately enable lower-field MR applications.   
%and labor costs and increasing the accessibility of MRI. Additionally, since it doesn't require high-fidelity images, $k$-space data collected from low-field MRI machines or portable MRI machines are potentially implementable with this method. 
%We do not envision our approach to replace conventional diagnostic procedures.
%is important to note that our method is not intended to replace conventional diagnostic procedures, as we recognise that image interpretation 
%by professional radiologists is essential for accurate biomarker localisation and diagnosis. 
We envision applications in pre-screening by offering preliminary diagnostic results to aid in resource allocation, particularly in regions with limited medical resources.
%thus avoiding unnecessarily lengthy diagnostic times.        

\noindent \textbf{Limitations}: Our proof of concept is based on the fastMRI knee data and specifically MT. One could readily see the application of a portable low-field scanner in sports medicine~\cite{massimiliano2019role} but it is a less compelling clinical application. Ideally, a dataset of e.g.\ stroke and acute brain trauma~\cite{lyu2023m4raw}, which would probe diverse, and not only structural, MR signal characteristics would be more compelling. At this moment such dataset is currently lacking. Our policy network relies on a pre-trained classifier, which requires pre-training with diverse undersampled data. It is of great interest to train the classifier simultaneously with the policy network. Finally, our policy network does not leverage physical priors of $k$-space utility which are implicitly leveraged by reconstruction policies using a pre-trained reconstruction network. Such prior also may encode physical limitations~\cite{kuperman2000magnetic} of the employed sequence in making sampling decisions. 

\section{Conclusion} 
Our proposed framework presents a novel approach to enhance the efficiency and accessibility of MRI as a Point of Care (POC) diagnostic tool by optimising sampling patterns in a learnable fashion. Our approach distinctly stipulates that $k$-space sampling acquisition decisions are undertaken by an agent optimising a dynamic per-patient classification task. Results reporting direct inference from undersampled $k$-space data, concerning  
%By incorporating active sampling policy into the diagnostic process, we demonstrated the feasibility of 
%inferring 
the presence of Meniscus Tear, showed that our approach achieves the comparable performance as an \textit{Oracle} Classifier whilst reducing acquisition time with only 25\% $k$-space usage. Furthermore, our analysis of task-specific sampling policies revealed that the policy network adapted its sampling strategy based on the nature of the task, demonstrating the adaptability of the active sampling approach. 
%Additionally, we acknowledge the method is designed for initial diagnosis and should complement rather than replace conventional diagnostic procedures.  

% \section*{Acknowledgements}
% This work was supported in part by National Institutes of Health (NIH) grant 7R01HL148788-03. Y. Du thanks additional financial support from the School of Engineering, the University of Edinburgh. 
% S.A.\ Tsaftaris also acknowledges the support of Canon Medical and the Royal Academy of Engineering and the Research Chairs and Senior Research Fellowships scheme (grant RCSRF1819\textbackslash 8\textbackslash 25), and the UK’s Engineering and Physical Sciences Research Council (EPSRC) support via grant EP/X017680/1.  The authors would like to thank Dr. Chen and K. Vilouras for inspirational discussions and assistance. 
\newpage
\bibliographystyle{splncs04} % Specify the bibliography style
\bibliography{reference} % Mention the BibTeX file name

\begin{thebibliography}{10}
\providecommand{\url}[1]{\texttt{#1}}
\providecommand{\urlprefix}{URL }
\providecommand{\doi}[1]{https://doi.org/#1}

\bibitem{bahadir2020deep}
Bahadir, C.D., Wang, A.Q., Dalca, A.V., Sabuncu, M.R.: Deep-learning-based optimization of the under-sampling pattern in mri. IEEE Transactions on Computational Imaging  \textbf{6},  1139--1152 (2020)

\bibitem{bakker_experimental_2020}
Bakker, T., van Hoof, H., Welling, M.: Experimental design for mri by greedy policy search. Advances in Neural Information Processing Systems  \textbf{33},  18954--18966 (2020)

\bibitem{cai2020review}
Cai, L., Gao, J., Zhao, D.: A review of the application of deep learning in medical image classification and segmentation. Annals of translational medicine  \textbf{8}(11) (2020)

\bibitem{geethanath2019accessible}
Geethanath, S., Vaughan~Jr, J.T.: Accessible magnetic resonance imaging: a review. Journal of Magnetic Resonance Imaging  \textbf{49}(7),  e65--e77 (2019)

\bibitem{he2016identity}
He, K., Zhang, X., Ren, S., Sun, J.: Identity mappings in deep residual networks. In: Computer Vision--ECCV 2016: 14th European Conference, Amsterdam, The Netherlands, October 11--14, 2016, Proceedings, Part IV 14. pp. 630--645. Springer (2016)

\bibitem{kool2019buy}
Kool, W., van Hoof, H., Welling, M.: Buy 4 reinforce samples, get a baseline for free!  (2019)

\bibitem{kuperman2000magnetic}
Kuperman, V.: Magnetic resonance imaging: physical principles and applications. Elsevier (2000)

\bibitem{lin2021artificial}
Lin, D.J., Johnson, P.M., Knoll, F., Lui, Y.W.: Artificial intelligence for mr image reconstruction: an overview for clinicians. Journal of Magnetic Resonance Imaging  \textbf{53}(4),  1015--1028 (2021)

\bibitem{lyu2023m4raw}
Lyu, M., Mei, L., Huang, S., Liu, S., Li, Y., Yang, K., Liu, Y., Dong, Y., Dong, L., Wu, E.X.: M4raw: A multi-contrast, multi-repetition, multi-channel mri k-space dataset for low-field mri research. Scientific Data  \textbf{10}(1), ~264 (2023)

\bibitem{massimiliano2019role}
Massimiliano, L., Giuseppe, G., Michela, B., Michele, A., Silvia, R., Alessio, P., Francesco, P., Lia, R., Alessandro, S., Federico, A.G., et~al.: Role of low field mri in detecting knee lesions. Acta Bio Medica: Atenei Parmensis  \textbf{90}(Suppl 1), ~116 (2019)

\bibitem{pineda_active_2020}
Pineda, L., Basu, S., Romero, A., Calandra, R., Drozdzal, M.: Active mr k-space sampling with reinforcement learning. In: Medical Image Computing and Computer Assisted Intervention--MICCAI 2020: 23rd International Conference, Lima, Peru, October 4--8, 2020, Proceedings, Part II 23. pp. 23--33. Springer (2020)

\bibitem{ravula_optimizing_2023}
Ravula, S., Levac, B., Jalal, A., Tamir, J.I., Dimakis, A.G.: Optimizing sampling patterns for compressed sensing mri with diffusion generative models. arXiv preprint arXiv:2306.03284  (2023)

\bibitem{ronneberger2015u}
Ronneberger, O., Fischer, P., Brox, T.: U-net: Convolutional networks for biomedical image segmentation. In: Medical Image Computing and Computer-Assisted Intervention--MICCAI 2015: 18th International Conference, Munich, Germany, October 5-9, 2015, Proceedings, Part III 18. pp. 234--241. Springer (2015)

\bibitem{schlemper2018cardiac}
Schlemper, J., Oktay, O., Bai, W., Castro, D.C., Duan, J., Qin, C., Hajnal, J.V., Rueckert, D.: Cardiac mr segmentation from undersampled k-space using deep latent representation learning. In: Medical Image Computing and Computer Assisted Intervention--MICCAI 2018: 21st International Conference, Granada, Spain, September 16-20, 2018, Proceedings, Part I. pp. 259--267. Springer (2018)

\bibitem{singhal_feasibility_2023}
Singhal, R., Sudarshan, M., Mahishi, A., Kaushik, S., Ginocchio, L., Tong, A., Chandarana, H., Sodickson, D.K., Ranganath, R., Chopra, S.: On the feasibility of machine learning augmented magnetic resonance for point-of-care identification of disease. arXiv preprint arXiv:2301.11962  (2023)

\bibitem{sondik1971optimal}
Sondik, E.J.: The optimal control of partially observable Markov processes. Stanford University (1971)

\bibitem{wang_joint_2022}
Wang, J., Yang, Q., Yang, Q., Xu, L., Cai, C., Cai, S.: Joint optimization of cartesian sampling patterns and reconstruction for single-contrast and multi-contrast fast magnetic resonance imaging. Computer Methods and Programs in Biomedicine  \textbf{226},  107150 (2022)

\bibitem{xuan_learning_2020}
Xuan, K., Sun, S., Xue, Z., Wang, Q., Liao, S.: Learning mri k-space subsampling pattern using progressive weight pruning. In: Medical Image Computing and Computer Assisted Intervention--MICCAI 2020: 23rd International Conference, Lima, Peru, October 4--8, 2020, Proceedings, Part II 23. pp. 178--187. Springer (2020)

\bibitem{zbontar2018fastmri}
Zbontar, J., Knoll, F., Sriram, A., Murrell, T., Huang, Z., Muckley, M.J., Defazio, A., Stern, R., Johnson, P., Bruno, M., et~al.: fastmri: An open dataset and benchmarks for accelerated mri. arXiv preprint arXiv:1811.08839  (2018)

\bibitem{zhang_reducing_2019}
Zhang, Z., Romero, A., Muckley, M.J., Vincent, P., Yang, L., Drozdzal, M.: Reducing uncertainty in undersampled mri reconstruction with active acquisition. In: Proceedings of the IEEE/CVF Conference on Computer Vision and Pattern Recognition. pp. 2049--2058 (2019)

\bibitem{zhao2021fastmri+}
Zhao, R., Yaman, B., Zhang, Y., Stewart, R., Dixon, A., Knoll, F., Huang, Z., Lui, Y.W., Hansen, M.S., Lungren, M.P.: fastmri+: Clinical pathology annotations for knee and brain fully sampled multi-coil mri data. arXiv preprint arXiv:2109.03812  (2021)

\end{thebibliography}

\newpage
\section*{Supplementary}

% \begin{figure}
%     \centering
%     \includegraphics[width=1\linewidth]{figs/metrics_cf.pdf}
%     \caption{Difference in Classification Performance with Various Center Fraction.}
%     \label{fig:enter-label}
% \end{figure}

\begin{table}
\centering
\caption{Image Classification Performance with Various Sample Rates and Center Fraction. Numbers in bold represent the upper bound.}
\resizebox{1.0\linewidth}{!}{
\begin{tabular}{c|cccccc|cccccc|cccccc}
\hline
 \textbf{Center Fraction}& \multicolumn{6}{|c}{\textbf{0.00}}& \multicolumn{6}{|c}{\textbf{0.01}}& \multicolumn{6}{|c}{\textbf{0.05}}\\  
\hline
\multicolumn{19}{l}{\textbf{AUC$\uparrow$}} \\
\hline
Sample Rate & 5\%& 7\%& 10\%& 13\%& 17\%& 25\%
& 5\%& 7\%& 10\%& 13\%& 17\%& 25\%
& 5\%& 7\%& 10\%& 13\%& 17\%& 25\% \\
\hline
Oracle &\textbf{0.819} & \textbf{0.819}&\textbf{ 0.819} & \textbf{0.819} & \textbf{0.819} &\textbf{ 0.819} & &\textbf{ 0.819} & \textbf{0.819} & \textbf{0.819} & \textbf{0.819} & \textbf{0.819} & \textbf{0.819} & \textbf{0.819} & \textbf{0.819} & \textbf{0.819} & \textbf{0.819} & \textbf{0.819} \\
Undersampled & 0.666 & 0.689 & 0.710 & 0.736 & 0.745 & 0.756 & 0.729 & 0.761 & 0.789 & 0.764 & 0.776 & 0.783 & 0.752 & 0.751 & 0.747 & 0.745 & 0.743 & 0.749 \\
Recon Classifier & 0.586 & 0.612 & 0.648 & 0.647 & 0.652 & 0.687 & 0.782 & 0.786 & 0.774 & 0.773 & 0.752 & 0.778 & 0.798 & 0.813 & 0.829 & 0.835 & 0.838 & 0.836 \\
Policy Reconstruction & 0.583& 0.806& 0.829& 0.844& 0.841& 0.837
& 0.757 & 0.808 & 0.817 & 0.826 & 0.827 & 0.813 & 0.789 & 0.818 & 0.833 & 0.829 & 0.827 & 0.831 \\
Policy Classifier & 0.666& 0.777& 0.775& 0.781& 0.785& 0.791
& 0.717 & 0.789 & 0.785 & 0.784 & 0.775 & 0.794 & 0.752 & 0.755 & 0.739 & 0.742 & 0.741 & 0.743 \\
\hline
% \multicolumn{19}{l}{} \\
% \hline
%  Acceleration Rate & 20 & 15 & 10 & 8 & 6 & 4 & 20 & 15 & 10 & 8 & 6 & 4 & 20 & 15 & 10 & 8 & 6 & 4 \\
%  \hline
% Oracle & \textbf{0.390} & \textbf{0.390} & \textbf{0.390} & \textbf{0.390} &\textbf{ 0.390 }& \textbf{0.390} & \textbf{0.390} & \textbf{0.390} & \textbf{0.390} & \textbf{0.390} & \textbf{0.390} & \textbf{0.390} & \textbf{0.390} & \textbf{0.390} & \textbf{0.390} & \textbf{0.390} & \textbf{0.390} & \textbf{0.390} \\
% Undersampled & 0.541 & 0.513 & 0.481 & 0.471 & 0.460 & 0.411 & 0.350 & 0.334 & 0.356 & 0.360 & 0.360 & 0.359 & 0.344 & 0.344 & 0.342 & 0.343 & 0.350 & 0.348 \\
% Recon Classifier & 0.438 & 0.440 & 0.406 & 0.401 & 0.429 & 0.402 & 0.487 & 0.487 & 0.537 & 0.545 & 0.568 & 0.551 & 0.349 & 0.360 & 0.387 & 0.404 & 0.436 & 0.448 \\
% Policy Reconstruction & 0.538& 0.557& 0.542& 0.520& 0.509& 0.510
% & 0.467 & 0.526 & 0.503 & 0.513 & 0.519 & 0.515 & 0.346 & 0.390 & 0.416 & 0.433 & 0.425 & 0.426 \\
% Policy Classifier & 0.541& 0.360& 0.350& 0.348& 0.349& 0.359
% & 0.337 & 0.331 & 0.340 & 0.342 & 0.346 & 0.351 & 0.344 & 0.346 & 0.350 & 0.352 & 0.352 & 0.356 \\
% \hline
\multicolumn{19}{l}{\textbf{RECALL$\uparrow$}} \\
\hline
Sample Rate & 5\%& 7\%& 10\%& 13\%& 17\%& 25\%
& 5\%& 7\%& 10\%& 13\%& 17\%& 25\%
& 5\%& 7\%& 10\%& 13\%& 17\%& 25\% \\
 \hline
Oracle & \textbf{0.816} & \textbf{0.816} & \textbf{0.816} & \textbf{0.816} & \textbf{0.816} & \textbf{0.816} & \textbf{0.816} & \textbf{0.816} & \textbf{0.816} & \textbf{0.816} & \textbf{0.816} & \textbf{0.816} & \textbf{0.816} & \textbf{0.816} & \textbf{0.816} & \textbf{0.816} & \textbf{0.816} &\textbf{0.816} \\
Undersampled & 0.592 & 0.636 & 0.655 & 0.699 & 0.699 & 0.694 & 0.592 & 0.655 & 0.728 & 0.680 & 0.704 & 0.723 & 0.636 & 0.636 & 0.626 & 0.621 & 0.621 & 0.641 \\
Recon Classifier & 0.243 & 0.311 & 0.388 & 0.379 & 0.408 & 0.510 & 0.762 & 0.772 & 0.767 & 0.762 & 0.733 & 0.791 & 0.748 & 0.791 & 0.840 & 0.854 & 0.869 & 0.869 \\
Policy Reconstruction & 0.228& 0.850& 0.893& 0.917& 0.908& 0.888
& 0.699 & 0.835 & 0.859 & 0.884 & 0.888 & 0.854 & 0.728 & 0.816 & 0.859 & 0.854 & 0.845 & 0.854 \\
Policy Classifier & 0.592& 0.699& 0.694& 0.709& 0.718& 0.738
& 0.544 & 0.714 & 0.709 & 0.709 & 0.694 & 0.709 & 0.636 & 0.646 & 0.617 & 0.621 & 0.621 & 0.626 \\
\hline
\multicolumn{19}{l}{\textbf{SPECIFICITY$\uparrow$}} \\
\hline
Sample Rate & 5\%& 7\%& 10\%& 13\%& 17\%& 25\%
& 5\%& 7\%& 10\%& 13\%& 17\%& 25\%
& 5\%& 7\%& 10\%& 13\%& 17\%& 25\% \\
 \hline
Oracle & \textbf{0.822} &\textbf{ 0.822} &\textbf{ 0.822} & \textbf{0.822} & \textbf{0.822} & \textbf{0.822} & \textbf{0.822} & \textbf{0.822} & \textbf{0.822} & \textbf{0.822} & \textbf{0.822} & \textbf{0.822} & \textbf{0.822} & \textbf{0.822} & \textbf{0.822} & \textbf{0.822} & \textbf{0.822} & \textbf{0.822} \\
Undersampled & 0.740 & 0.741 & 0.765 & 0.774 & 0.791 & 0.817 & 0.866 & 0.866 & 0.851 & 0.848 & 0.848 & 0.843 & 0.869 & 0.865 & 0.869 & 0.869 & 0.864 & 0.857 \\
Recon Classifier & 0.928 & 0.914 & 0.907 & 0.916 & 0.895 & 0.864 & 0.801 & 0.799 & 0.781 & 0.783 & 0.771 & 0.764 & 0.849 & 0.834 & 0.818 & 0.815 & 0.807 & 0.804 \\
Policy Reconstruction & 0.938& 0.763& 0.765& 0.770& 0.774& 0.785
& 0.815 & 0.780 & 0.774 & 0.768 & 0.766 & 0.771 & 0.850 & 0.821 & 0.807 & 0.804 & 0.810 & 0.808 \\
Policy Classifier & 0.740& 0.854& 0.856& 0.853& 0.851& 0.843
& 0.890 & 0.864 & 0.861 & 0.860 & 0.856 & 0.852 & 0.869 & 0.864 & 0.862 & 0.862 & 0.861 & 0.861 \\
\hline

\end{tabular}}

\end{table}

\begin{figure}
    \centering
    \includegraphics[width=1\linewidth]{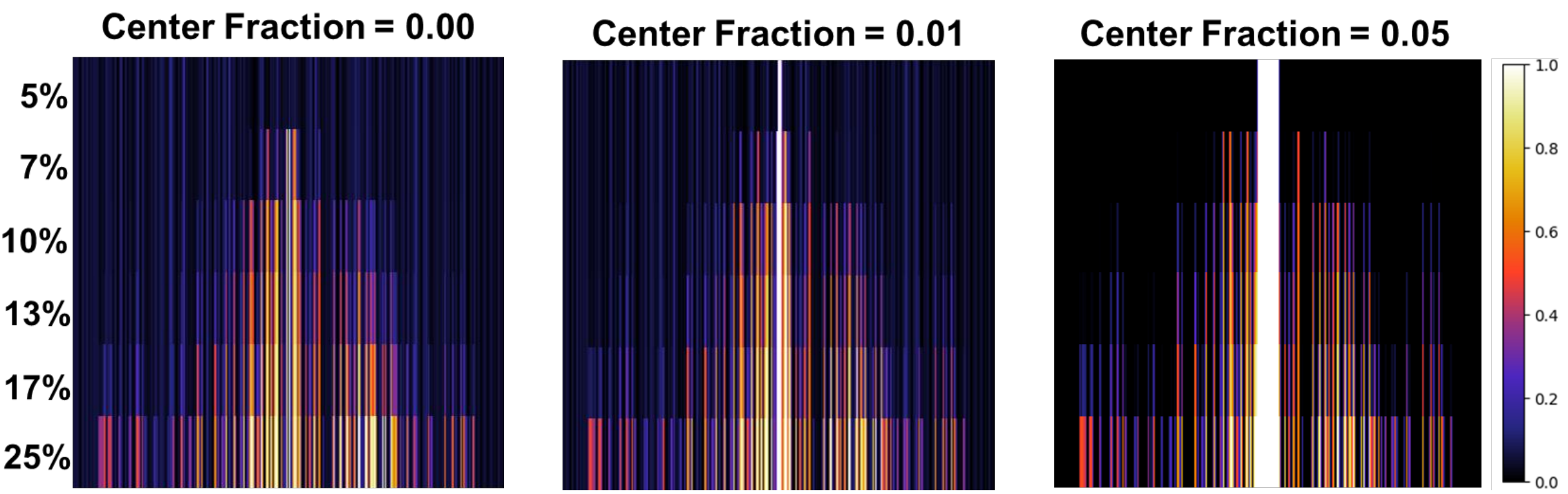}
    \caption{$k$-space preference for reconstruction task as a function of sample rate (vertical axis downward) and for 3 center fraction scenarios. Colorbar indicates the possibility of being sampled.}
    \label{fig:enter-label}
\end{figure}

\end{document}